%% file: fair_autoencoder.tex
\documentclass[runningheads]{llncs}
\usepackage{pgfplots}
\pgfplotsset{width=15cm,compat=1.8}
\usepackage{pgfplotstable}
\usepackage{graphicx}
\usepackage{subfigure}
\usepackage{booktabs}
\usepackage{pgfplots}
\usepackage{amsmath}

\usepackage[pagebackref=true,breaklinks=true,letterpaper=true,colorlinks,bookmarks=false]{hyperref}

\newcommand\ourmethod{\textit{FairNN}~}

\begin{document}
\title{\ourmethod ~- Conjoint Learning of Fair Representations for Fair Decisions
}
%
%
\author{Tongxin Hu\inst{1} \and
Vasileios Iosifidis\inst{2} \and
Wentong Liao\inst{1} \and
Hang Zhang\inst{1} \and
Michael Ying Yang\inst{3}\and
Eirini Ntoutsi\inst{2} \and
Bodo Rosenhahn\inst{1}}

%
%
\institute{$^1$Institut f\"ur Informationsverarbeitung, Leibniz University of Hanover, Germany\\
$^2$L3S Research Center, Leibniz University of Hanover, Germany\\
$^3$Scene Understanding Group, University of Twente, Netherlands\\
$^1$\email{\{second\_name\}@tnt.uni-hannover.de}, $^2$\email{\{second\_name\}@l3s.de}, $^3$\email{michael.yang@utwente.nl}}

\maketitle              
\begin{abstract}

In this paper, we propose \ourmethod a neural network that performs joint feature representation and classification for fairness-aware learning.
Our approach optimizes a multi-objective loss function in which  (a) learns a fair representation  by suppressing protected attributes   (b) maintains the information content  by minimizing a reconstruction loss and  (c) allows for solving a classification task in a fair manner  by minimizing the classification error and respecting the equalized odds-based fairness regularizer.
Our experiments on a variety of datasets demonstrate that such a joint approach is superior  to separate treatment of unfairness in  representation learning or supervised learning. Additionally, our regularizers can be adaptively weighted to balance the different components of the loss function, thus allowing for a very general framework for conjoint fair representation learning and decision making.


\end{abstract}
%
%
\section{Introduction}
\label{sec:intro}
\input{introduction}
\section{Related work}
\label{sec:related}
\input{related}

\section{Basic concepts and definitions}
\label{sec:basics}
\input{basics}



\section{\ourmethod}
\label{sec:method}
\begin{figure}[t!]
	\includegraphics[width=1.05\textwidth]{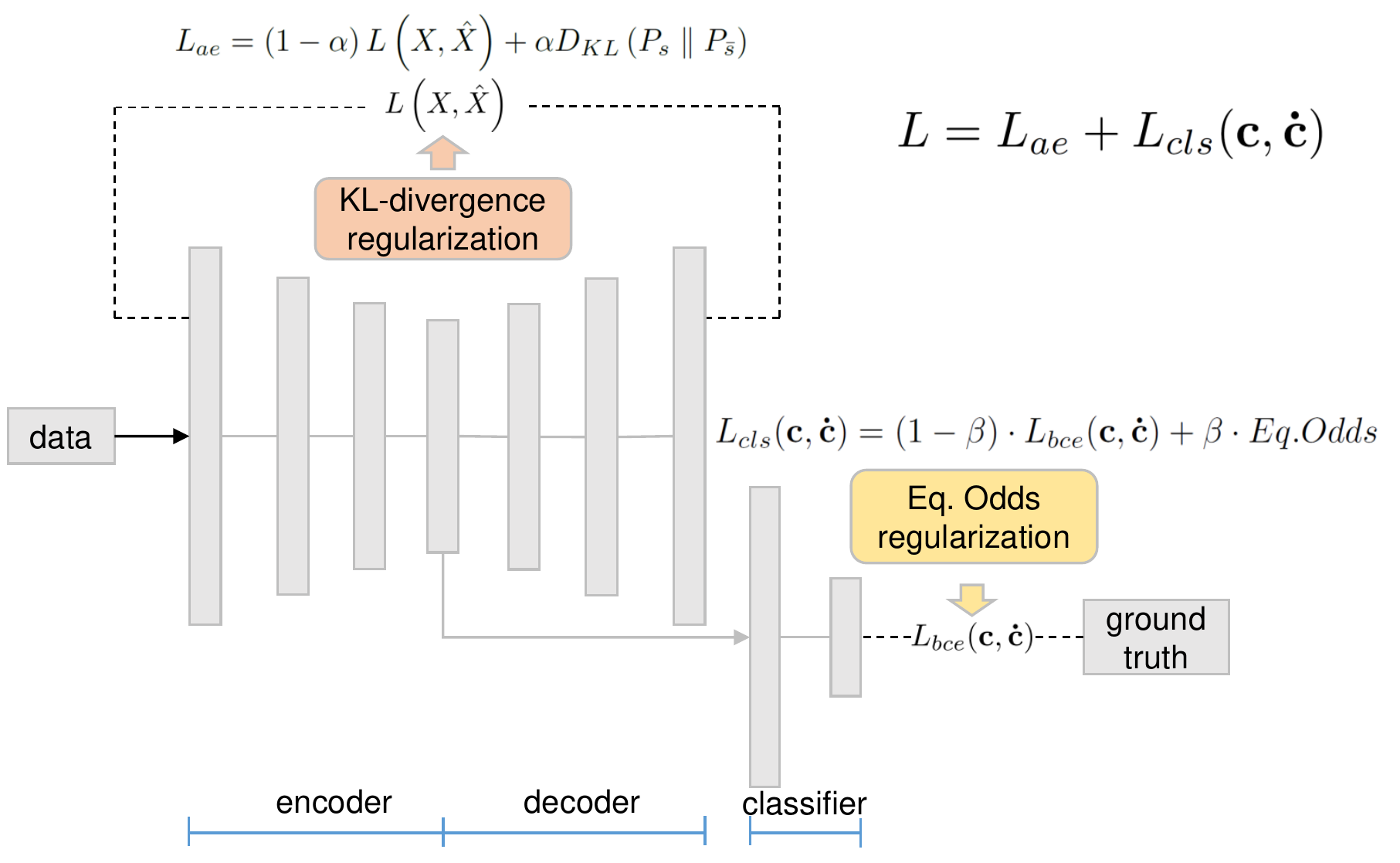}
	\caption{An overview of \ourmethod that jointly learns a fair representation and a fair mapping function for classification. The auto-encoder (left part) is responsible for representation learning; the KL-divergence constraint forces the representation to be fair. The loss function of the classifier (right part) is tweaked towards fairness through the Eq.Odds regularization. Both aspects are reflected in the joint objective  
	} \label{fig:framework}
\end{figure}

In this section, we introduce our proposed method, namely~\ourmethod that jointly learns a fair representation and a fair mapping function for classification. An overview of our approach is depicted in Fig.~\ref{fig:framework}\footnote{Source code will be made publicly available upon acceptance.}.
The architecture consists of two parts, an auto-encoder block aiming at learning a fair latent representation of the data (left) and a classification block aiming at learning a fair classifier (right).
We explicitly consider fairness in the representation learning by adding an additional constraint to the latent space of the auto-encoder in order to obfuscate the information on the protected attribute
(Section~\ref{subsec:kld}). 
Likewise, we explicitly consider fairness in the classification part by adding an additional constraint to the loss function based on the Equalized Odds fairness notion (Eq.~\ref{eq:EqOdds}) (Section~\ref{subsec:eo}). 
We consider these aspects \emph{jointly} and optimize a multi-loss objective function that balances the importance of the different components in-training (Section~\ref{sec:multiloss}).

\subsection{Fair representation learning via KL-divergence regularization}
\label{subsec:kld}
In order to learn fair feature transformations for the protected and non-protected groups,  \emph{KL divergence} is added to the loss function to train the auto-encoder, which constrains the learned features of different groups to have similar distribution properties.
With this constraint, the auto-encoder is trained to mix up the protected attribute information and meanwhile to maintain good reconstruction ability. 
In practice, we use the \emph{KL divergence} as an additional regularization in the objective function.
Based on the values of protected attributes, we divide the data points into protected group $s$ and non-protected group $\bar{s}$. 
Without loss of generality, we assume their distribution in the latent space as \emph{$d$-dimensional normal distributions}
with means $\mu_{s}$, $\mu_{\bar{s}}$ and covariance matrices $\Sigma_{s}$, $\Sigma_{\bar{s}}$ respectively. 
Then, the \textit{KL divergence} between the their distributions is given as:
\begin{equation}
D_{KL}\left  ( P_{s}\parallel P_{\bar{s}} \right )=\frac{1}{2}\left  ( \log\frac{det\left  ( \Sigma _{\bar{s}} \right )}{det\left  ( \Sigma _{s} \right )}-d+tr\left  ( \Sigma_{\bar{s}}^{-1}\Sigma _{s} \right )+\left  ( \mu _{\bar{s}}-\mu_{s} \right )^{T}\Sigma _{\bar{s}}^{-1}\left  ( \mu _{\bar{s}}-\mu_{s}\right ) \right )
\label{eq:kld}
\end{equation} 
where, $det (\Sigma)$ is the determinant of the covariance matrix $\Sigma$, and $tr (\cdot)$ is the trace of the matrix, which is the sum of elements on the main diagonal of the matrix.
With the \emph{KL-Divergence} Regularization, the original reconstruction loss function of the auto-encoder (c.f., Eq.~\ref{eq:multi_loss}) is rewritten as:
\begin{equation}
L_{ae}=\left  ( 1-\alpha \right ) L\left  ( X,\hat{X} \right ) + \alpha D_{KL}\left  ( P_{s}\parallel P_{\bar{s}} \right )	
\label{eq:ae_loss}
\end{equation} 
where $\alpha \in \left [ 0, 1 \right )$, is a coefficient for balancing the two terms.

\begin{figure}[!t]
	\subfigure[low-dimensional embedding without \textit{KL-Divergence} Regularizer]{
		\begin{minipage}{0.49\linewidth}
			\centering\includegraphics[width=\linewidth]{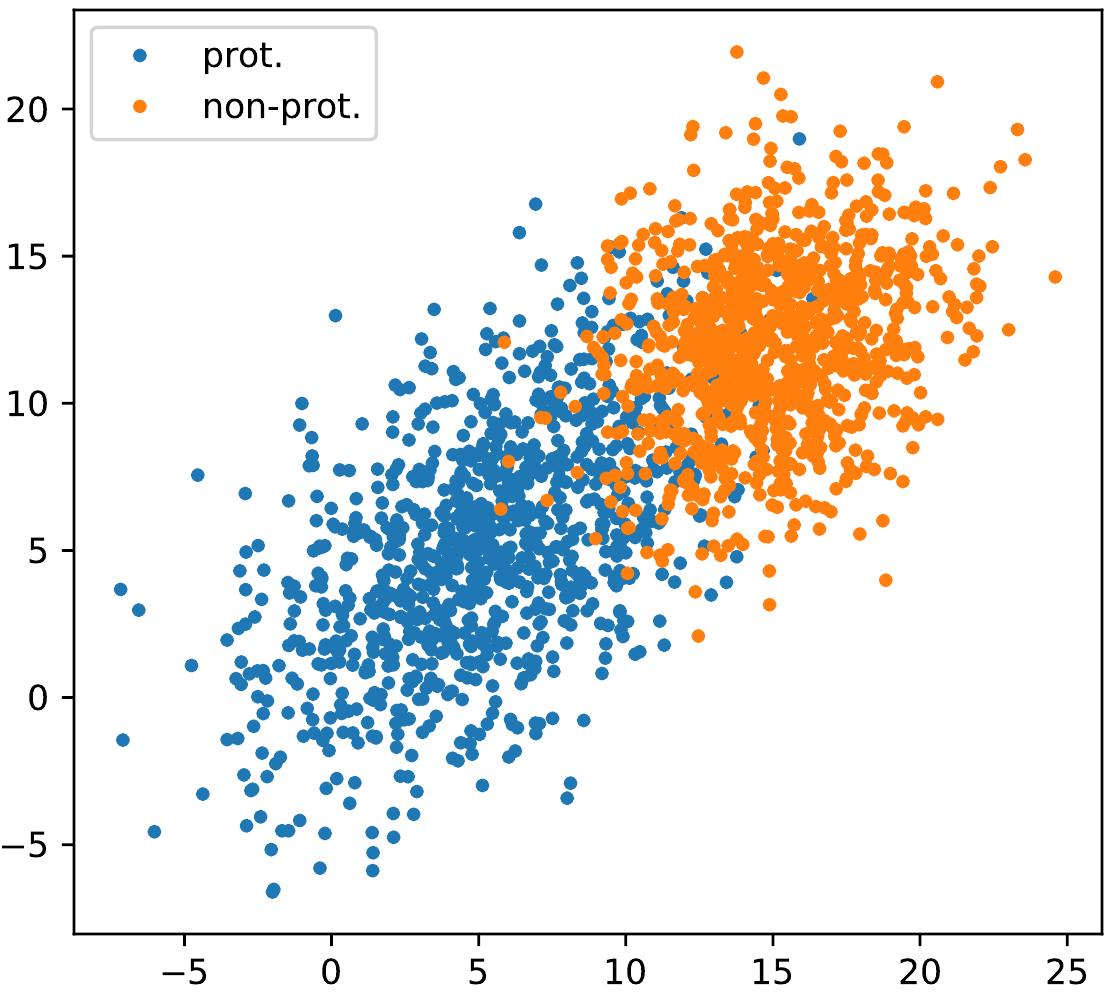}\\
		\end{minipage}
		\label{fig:embedding1}
	}
	\hfill
	\subfigure[low-dimensional embedding with \emph{KL-Divergence} Regularizer]{
		\begin{minipage}{0.5\linewidth}
			\centering\includegraphics[width=\linewidth]{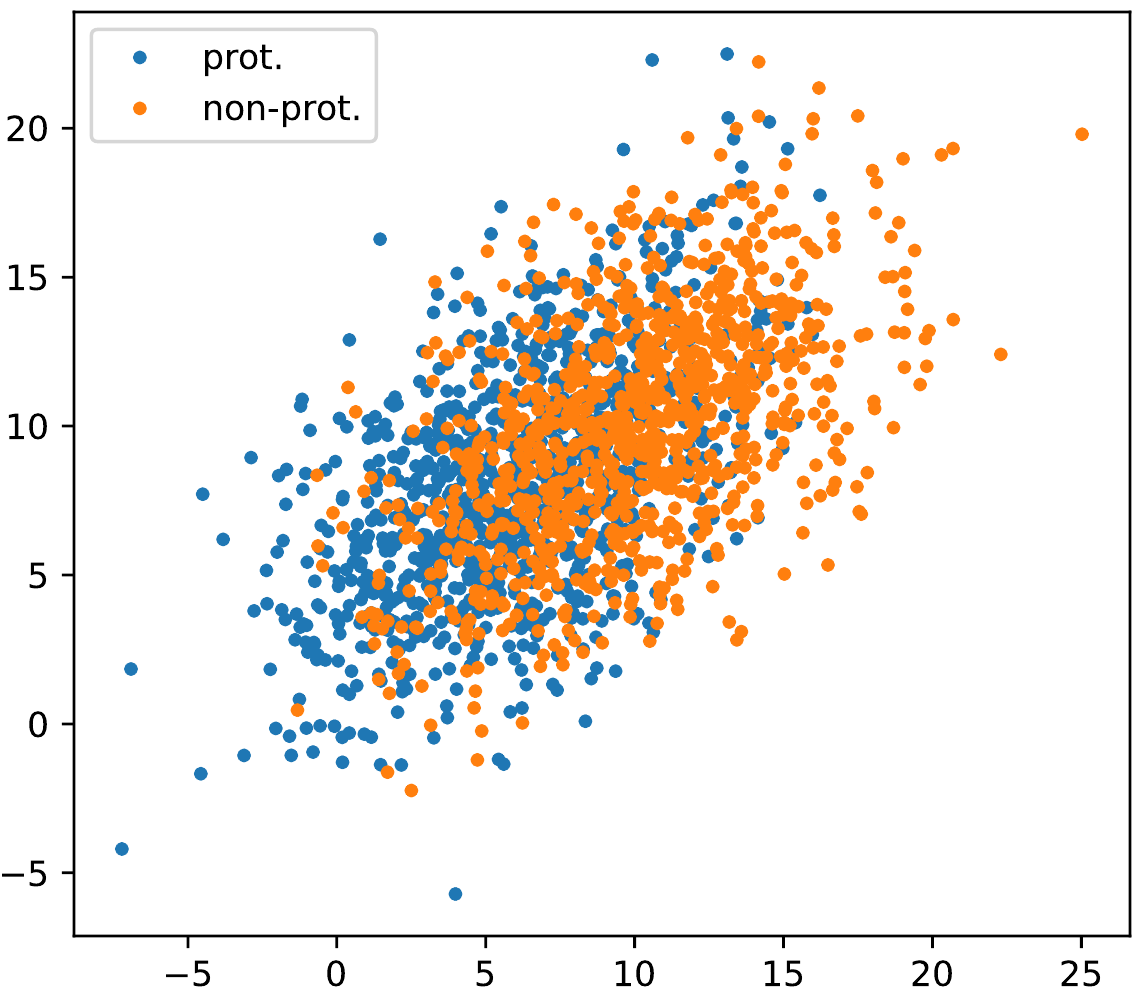}\\
		\end{minipage}
		\label{fig:embedding2}
	}
	\caption{Effect of the \emph{KL-Divergence} Regularizer in (fair) representation learning}
	\label{fig:embedding}
\end{figure}

Fig.~\ref{fig:embedding} demonstrates the impact of our KL-divergence regularizer, as distribution of data points in a low-dimensional feature space, in contrast to a transformation that has been learned without KL-Divergence regularization.
The protected and non-protected groups are denoted in blue and orange respectively.
 Fig.~\ref{fig:embedding1} shows that the data points belonging to different groups are easy to be separated in the latent space with direct implications to fairness. The regularizer mixes-up the distributions of the two groups making it hard to predict the protected attribute, c.f., Fig.~\ref{fig:embedding2}. 

\subsection{Fair classifier learning via equalized odds regularization}
\label{subsec:eo}
The classifier is an MLP with two FC layers followed by \emph{Relu} activation. The output is a scalar that is squashed by the sigmoid function between 0 and 1 for our binary classification task. The \emph{Binary Cross Entropy} is used as loss function to train the classifier as follows:
\begin{equation}
	L_{bce} (\mathbf{c},\mathbf{\dot{c}})=-\frac{1}{B}\sum_{n=1}^{b}\left  (\left  ( c_{b}\log\left  ( \dot{c}_b \right )+\left  ( 1-c_{b} \right )\log\left  (  1-\dot{c}_b\right ) \right )  \right )
	\label{eq:bce}
\end{equation} 
where $c_{b}$ is the true label and $\dot{c}_{b}$ is the predicted probability of the data point $b$ having the label $c_{b}$.

Our goal is to improve the fairness performance without losing the classification performance. This motivates us to add an additional  fairness measurement as a regularization term in the objective function. As we mentioned before, among different fairness measurements, \textit{Equalized Odds} does not only consider the predicted outcome but also compares it to the actual outcome recorded in the dataset. It considers both the samples with actual positive labels and also those with negative labels. Therefore, \emph{Equalized Odds}  (Eq.Odds) is used as the constraint term and added to the classification loss Eq.~\eqref{eq:bce}:
\begin{equation}
		L_{cls} (\mathbf{c},\mathbf{\dot{c}})=\left  ( 1-\beta \right )\cdot L_{bce} (\mathbf{c},\mathbf{\dot{c}})+\beta\cdot Eq. Odds
	\label{eq:mlp_loss}
\end{equation} 
where $\beta \in \left [ 0, 1 \right )$, is a balancing coefficient between the classification loss $L_{bce}$ and the \emph{Eq.Odds} fairness regularization.

\subsection{Fair Representation and Classifier-learning via Joint Optimization}
\label{sec:multiloss}
By combining the two parts of our network, which are the
auto-encoder  (Eq.~\ref{eq:ae_loss}) and classifier loss  (Eq.~\ref{eq:mlp_loss}), the acquired multi-loss function can be expressed as:
\begin{equation}
L = L_{ae} + L_{cls} (\mathbf{c},\mathbf{\dot{c}}).
\label{eq:loss_all}
\end{equation} 

It is known that neural networks can easily be over-parameterized and tend to overfit, given limited training data. The additional constraints in our architecture, together with the auto-encoder component enforces better generalization, as demonstrated in our experiments (Section~\ref{sec:experiment}).
We implemented  \ourmethod in the Python framework using PyTorch.


\section{Experiments}
\label{sec:experiment}
We evaluate the predictive and fairness performance of \ourmethod\footnote{Source code and data will be made available upon acceptance} and compare the results with recent state-of-the-art methods. Additionally, we perform several ablation studies to demonstrate the importance of each component in our proposed framework.
Accuracy and balanced accuracy are reported for evaluating the \emph{predictive} performance and \emph{Equalized Odds} for \emph{fairness} performance. Since \emph{Equalized Odds} reports the difference between two groups and we also want to maintain the predictive performance for both groups, we also report the actual \emph{TPR} and \emph{TNR} of both groups. 

\subsection{Experimental setup}
\label{subsec:setup}

\subsubsection{Datasets}
\label{subsubsec:dataset}
We evaluate our method on two real-world datasets, summarized in Tab.~\ref{tab:dataset}:
\begin{itemize}
    \item  \textbf{Adult Census Income Dataset}\cite{UCI2013} is extracted from the 1994 American Census Database. The task is to predict whether a person's income is over 50K a year. People with label $>$\textit{50K} belong to the positive class. $S$ = \emph{gender} is considered as the protected attribute, $s$ = \emph{female} the protected group and $\bar{s}$ = \emph{male} the non-protected group.
    \item  \textbf{Bank Marketing Dataset}\cite{moro2014data} is collected from a Portuguese bank that focuses on selling long-term deposits over the phone. The task is to predict whether a client will make a deposit subscription. We take $S$ = \emph{marital status} as the protected attribute, $s$ = \emph{married} the protected group and $\bar{s}$ = \emph{single/divorced} as the non-protected group.
\end{itemize}

\begin{table}[t!]
	\centering
	\caption{An overview of the datasets.}
	\resizebox{\textwidth}{!}{	
	\begin{tabular}{@{}ccccccc@{}}
		\toprule
		& \#Instances & \#Attributes & \begin{tabular}[c]{@{}c@{}}protected \\ attribute\end{tabular} & \begin{tabular}[c]{@{}c@{}}protected \\ group\end{tabular} & \begin{tabular}[c]{@{}c@{}}class \\ ratio (+:-)\end{tabular} & \begin{tabular}[c]{@{}c@{}}positive \\ class\end{tabular} \\ \midrule
		Adult Census                                                      & 45,175       & 14           & \textit{gender}                                                & \textit{female}                                            & 1:3.03                                                      & \textit{\textgreater{}50K}                                \\
		Bank Marketing                                                    & 40,004       & 16           & \textit{marital status}                                        & \textit{married}                                           & 1:7.57                                                      & \textit{yes}                                              \\
		\bottomrule
	\end{tabular}}
	\label{tab:dataset}
\end{table}

\subsubsection{Experimental settings}
\label{subsubsec:params}
The nominal attributes are encoded to one-hot vector and max-normalization is applied to the numerical attributes to ensure the values are in $[0, 1]$. In the auto-encoder block, both the encoder and decoder have three fully-connected linear layers and each is followed by a $ReLU$ activation. Following the evaluation setup in \cite{DBLP:conf/www/ZafarVGG17,DBLP:conf/www/KrasanakisXPK18,DBLP:conf/cikm/IosifidisN19}, $50\%$ of the data is used for training in which $20\%$ of them are used for validation, and the other $50\%$ is for testing. All experiments are evaluated using 10 random splits. We train the auto-encoder and classifier simultaneously by minimizing the objective function Eq.~\ref{eq:loss_all}. For training, we use the Adam optimization method, with batch size $B=512$ 
and a learning rate $0.002$. 
In order to get the best $\alpha-\beta$ combination (see Eq.~\eqref{eq:ae_loss}  and \ref{eq:mlp_loss}), grid search is operated within $\alpha \in [0.4, 0.5, 0.6, 0.7, 0.8, 0.9]$ and $\beta \in [0.1, 0.2, 0.3, 0.4, 0.5]$. Finally, $\alpha=0.9,\beta=0.2$ for the Adult Census Income Dataset and $\alpha=0.8, \beta=0.4$ for the Bank Marketing Dataset are selected.


\subsubsection{Preferential sampling}
\label{subsubsec:ps}

Due to the class imbalance problem, we leverage preferential sampling~\cite{DBLP:journals/kais/KamiranC11} to further improve the performance, which is a combination of oversampling the protected population and under-sampling the non-protected population. 
Note that, for the latter, we select those instances close to the decision boundary rather than random to avoid a big negative impact on the classifier.
At first, we make predictions with the learned classifier for all the samples in the training dataset, and then rank all the prediction scores. 
In $s_+$ and $\bar{s}_+$ we duplicate $k$ instances whereas in $s_{-}$ and $\bar{s}_{-}$ we remove $k$ instances near the classification boundary. $k$ is calculated with the formula $\frac{\left | v \right |\times \left | l \right |}{\left | D \right |}$, where $\left | v \right |$ is the number of instances with the community's corresponding sensitive value, $\left | l \right |$ the number of instances with the community's corresponding class label, and $\left | D \right |$ the size of the training dataset. At last, we continue training the network on this modified dataset.

\begin{figure}[!t]
	\subfigure[Comparison on Adult Census Income Dataset.]{
		\begin{minipage}{0.5\linewidth}
			\centering\includegraphics[width = 5.971cm, height = 6.223cm]{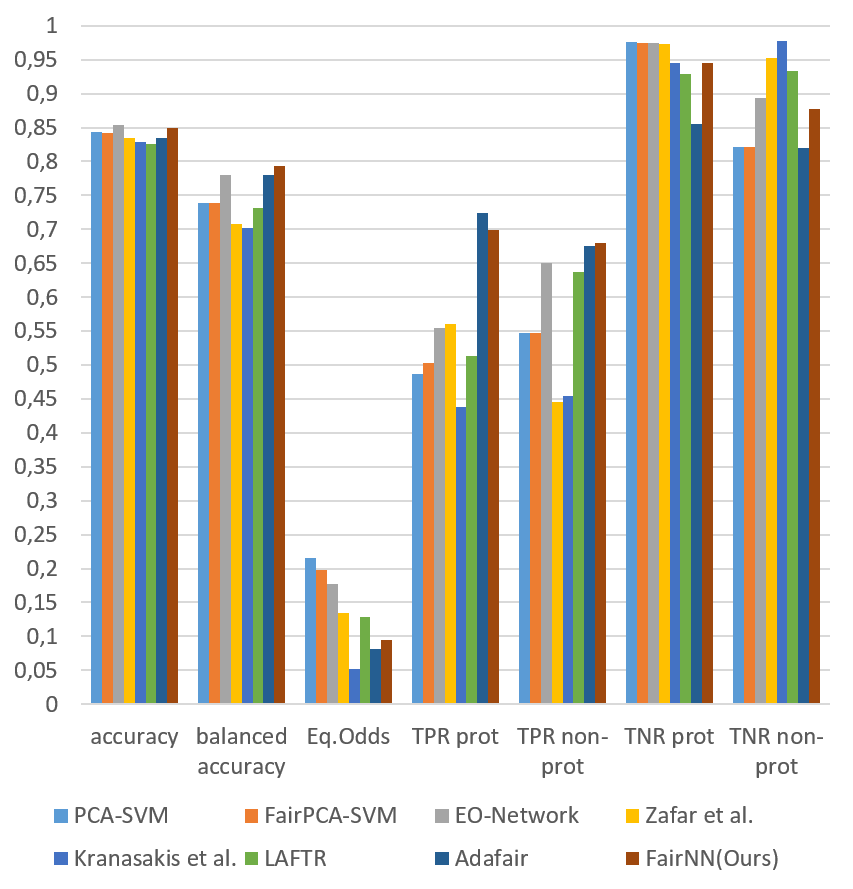}\\
		\end{minipage}
		\label{fig:com_adult}
	}
	\hfill
	\subfigure[Comparison on Bank Marketing Dataset. ]{
		\begin{minipage}{0.5\linewidth}
			\centering\includegraphics[width = 5.971cm, height = 6.223cm]{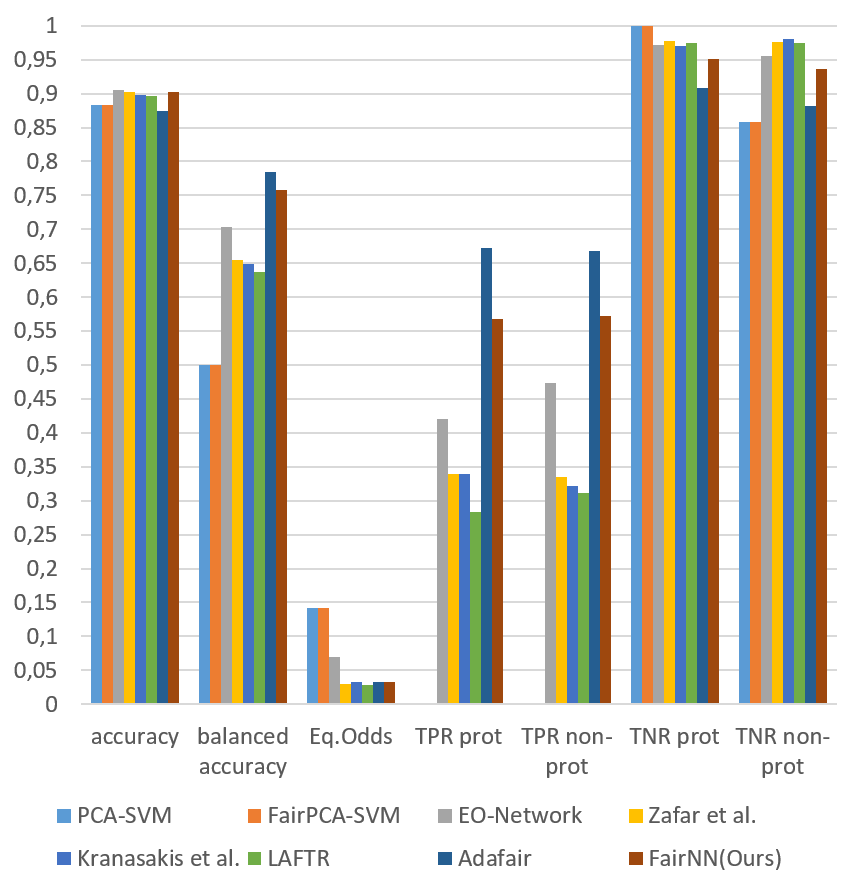}\\
		\end{minipage}
		\label{fig:com_bank}
	}
	\caption{Comparison with the state-of-the-art methods on Adult Census Income dataset and  Bank Marketing dataset. For fairness measurement \textit{Eq.Odds}, lower values are better; For others, higher are better.}
	\label{fig:comparison}
\end{figure}
\subsection{Comparison with other Methods}
\label{subsec:comparison}
We compared our approach with the recently proposed state-of-the-art in-processing approaches which mainly aim to minimize \emph{Eq.Odds}.
\begin{itemize}
    \item AdaFair~\cite{DBLP:conf/cikm/IosifidisN19}: a boosting model which assigns fairness related weights in each boosting round by observing the cumulative fairness behavior of the ensemble.
    \item LAFTR~\cite{madras2018learning}: a holistic approach that learns a latent fair representation using an encoder/decoder and an adversary  (where the encoder/decoder seek to minimize the adversary's objective), and at the same time trains a fair classifier on the latent space. 
    \item FairPCA-SVM~\cite{DBLP:conf/nips/SamadiTMSV18}: aims to find a low dimensional representation of the original data while maintaining similar fidelity for two groups. We project the data to the Fair PCA space and use SVM for binary classification.
    \item PCA-SVM: Similar to FairPCA-SVM, we project the data to the PCA space and use an SVM classifier. This is only a naive baseline method for comparison.
    \item EO-Network~\cite{manisha2018neural}: A two-layer neural network, with \textit{Eq.Odds} as a constraint in the loss function. This can be seen as our model without the auto-encoder part.
    \item Krasanakis et al.~\cite{DBLP:conf/www/KrasanakisXPK18}: In this work, the authors assume the existence of a latent fair class distribution, which they approximate through the CULEP model by re-estimating the instance weights iteratively.
    \item Zafar et al.~\cite{DBLP:conf/www/ZafarVGG17}: In this work, the authors formulate fairness as a set of convex-concave constrains which are embedded in the objective function of a logistic regression model.
\end{itemize}

The experimental results from different methods on two datasets are depicted in Fig.~\ref{fig:comparison}, detailed discussion on each dataset follows hereafter.
The results of \cite{DBLP:conf/www/KrasanakisXPK18,DBLP:conf/www/ZafarVGG17,DBLP:conf/cikm/IosifidisN19} are taken from \cite{DBLP:conf/cikm/IosifidisN19}. 

\begin{figure}[!t]
	\subfigure[Adult Census Dataset.]{
		\begin{minipage}{0.45\linewidth}
			\centering\includegraphics[width = 5.726cm, height = 6.223cm]{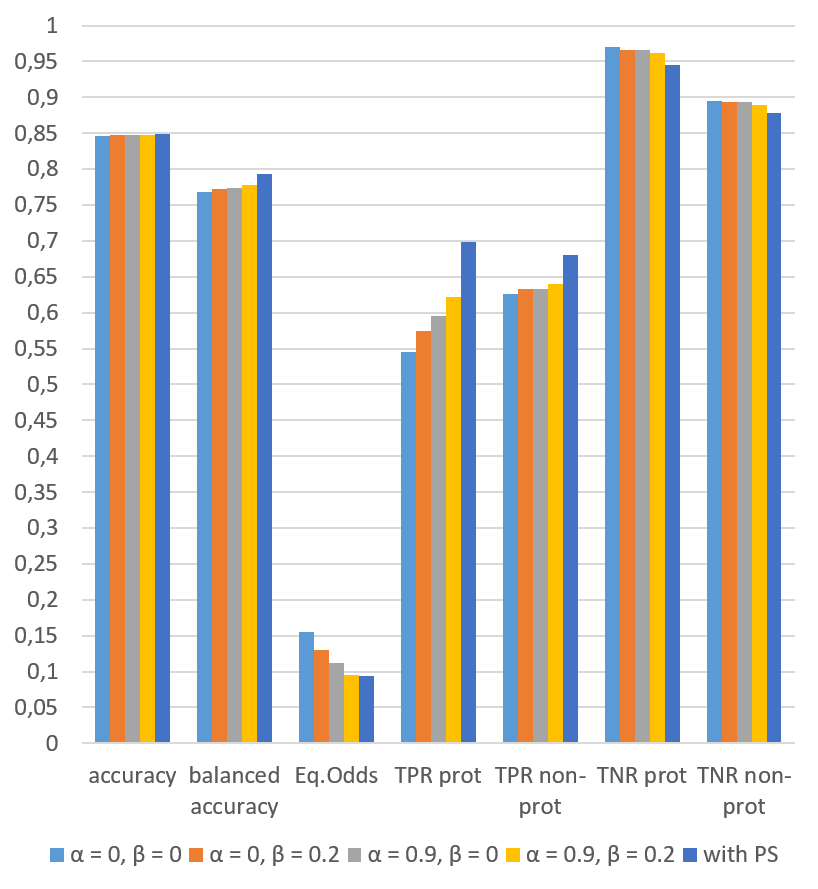}\\
		\end{minipage}
		\label{fig:ab_adult}
	}
	\hfill
	\subfigure[Bank Marketing Dataset.]{
		\begin{minipage}{0.5\linewidth}
			\centering\includegraphics[width = 5.726cm, height = 6.223cm]{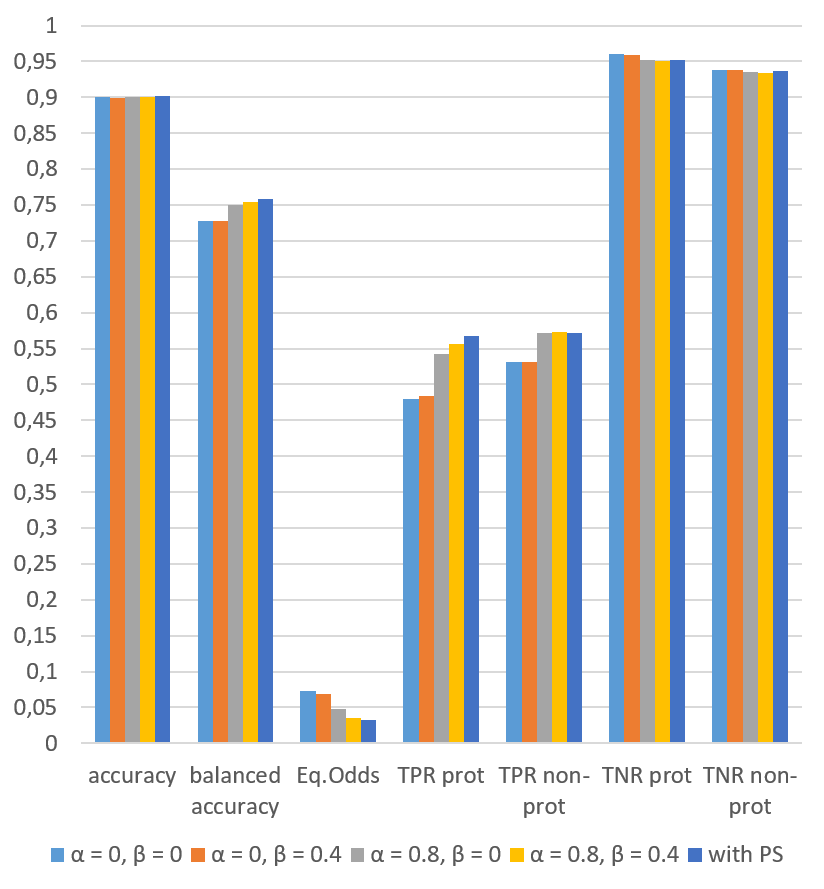}\\
		\end{minipage}
		\label{fig:ab_bank}
	}
	\caption{Ablation Study (for \textit{Eq.Odds} lower values are better - for the rest, higher values are better.)
	}
	\label{fig:ablation}
\end{figure}

\subsubsection{Adult Census Income}
\label{subsubsec:adult}
Fig.~\ref{fig:com_adult} displays the baselines, state-of-the-art and our final experimental results on the Adult Census Dataset. Our method achieves the highest accuracy and balanced accuracy rates. The lowest \textit{Eq.Odds} is achieved by Krasanakis et al. However, its TPRs for both \textit{protected}(TPR prot) and \textit{non-protected}(TPR non-prot) groups are much lower than the other methods (the lowest TPR prot and the second-lowest TPR non-prot). Fair-PCA aims to learn a fair feature representation in the low-dimensional space. But the learned representation may be unsuited for the binary classification task. It achieves fairer decision-making (lower \textit{Eq.Odds}) comparing to PCA yet performs worse compared to our method. The comparison of our method with EO-Network demonstrates an 8\% decrease in \textit{Eq.Odds} and 14\% improvement in TPR prot, revealing the effectiveness of generating low-dimensional features. Similar to our approach, LAFTR  also leverages the joint-learning thought, but ours is more effective comparing to theirs: balanced accuracy is 5\% higher and \textit{Eq.Odds} 3\% lower. Our method also brings a significant increase in TPR prot(18\% higher).
The superior performance from our method indicates that our method is able to learn the fair representation. It balances the balanced accuracy and \textit{Eq. Odds} well.

\subsubsection{Bank}
\label{subsubsec:bank}
In Fig.~\ref{fig:com_bank}, we report experimental results on the Bank Marketing Dataset. 
Due to the class imbalance problem, both PCA-SVM and FairPCA-SVM perform poorly on this dataset. They output all zeros for the binary classification task which result in balanced accuracy $0.5$, TPR prot and TPR non-prot are $0$. In EO-Network, the weight parameter of the \textit{Eq.Odds} constraint is the same as used in our method i.e., $0.4$. LAFTR reaches the lowest \textit{Eq. Odds} result but its TPRs for both groups are also the lowest. There is a minor difference in \textit{Eq.Odds} between LAFTR and our method, yet ours achieves the much higher balanced accuracy rate, TPR prot, and TPR non-prot. Compared to Zafar et al. and Krasanakis et al., our method reports a higher balanced accuracy rate, higher TPRs for both groups and also the comparable \textit{Eq.Odds}. It proves that our method maintains classification performance while achieving fairness. 

\subsection{Ablation study}
\label{subsec:ablation}
We perform ablation studies to evaluate how different parts influence the predictive and fairness performance of our method. In Fig.~\ref{fig:ablation}, $\alpha=0$ represents the outcome without \textit{KL-Divergence} regularization and $\beta=0$ without \textit{Eq.Odds} regularization respectively. Fig.~\ref{fig:ab_adult} demonstrates the ablation study on the Adult Census Income Dataset and Fig.~\ref{fig:ab_bank}the Bank Marketing Dataset. We can see that, integrating only the \textit{KL-Divergence} regularization is more effective than integrating \textit{Eq.Odds} regularization only  (comparing the second and third bars in Fig.~\ref{fig:ab_adult}).  Applying both regularizations further improves the performance  (the fourth bars in Fig.~\ref{fig:ab_adult}). Preferential sampling further improves the TPR prot and TPR non-prot while almost not affecting \textit{Eq.Odds}. The ablation study results on the Bank Dataset  (as shown in Fig.~\ref{fig:ab_bank}) display a similar tendency yet Preferential sampling does not bring much improvement on TPRs.

\begin{figure}[!t]
	\subfigure[colored by gender]{
		\begin{minipage}{0.5\linewidth}
			\centering\includegraphics[width=\linewidth]{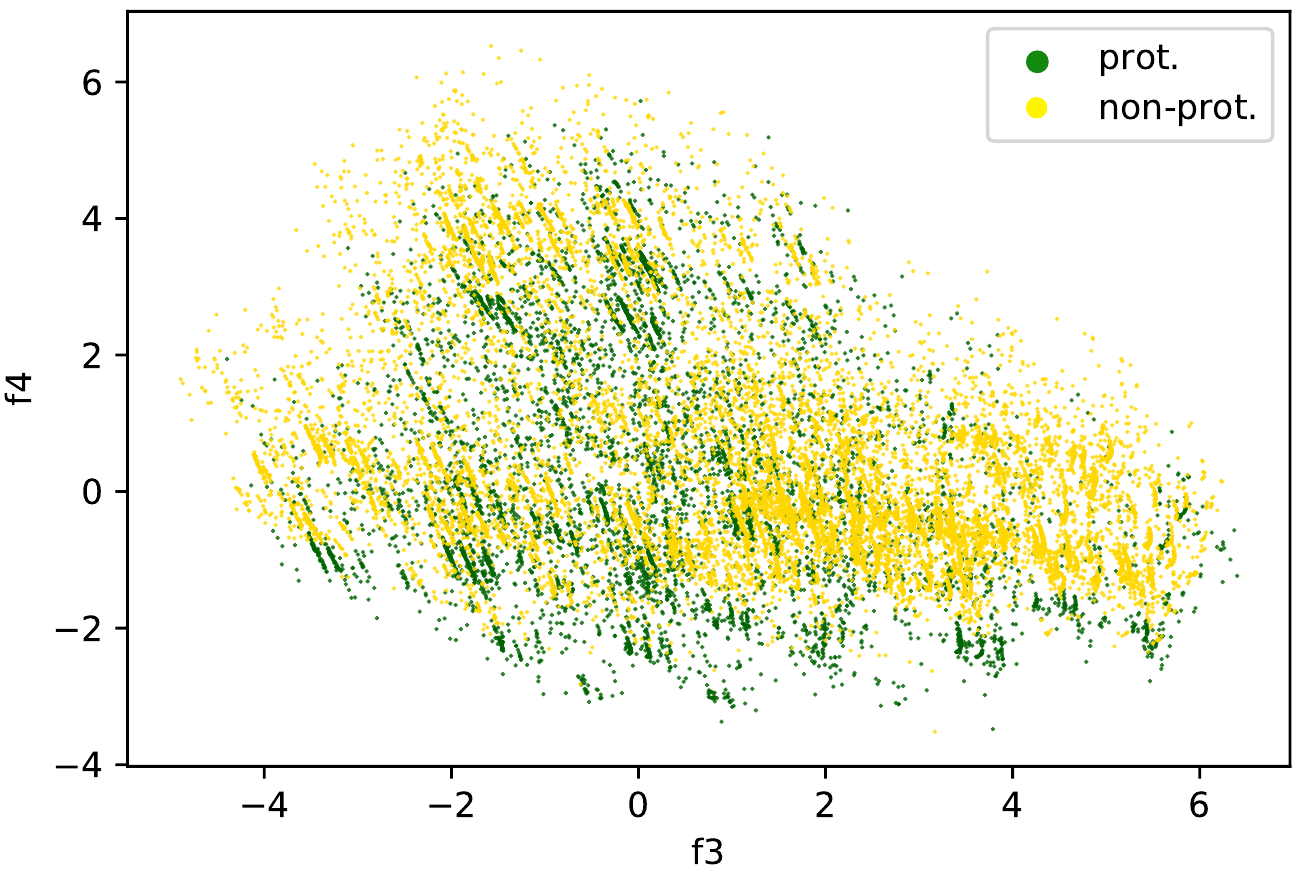}\\
		\end{minipage}
		\label{fig:visual_gender}
	}
	\hfill
	\subfigure[colored by income]{
		\begin{minipage}{0.5\linewidth}
			\centering\includegraphics[width=\linewidth]{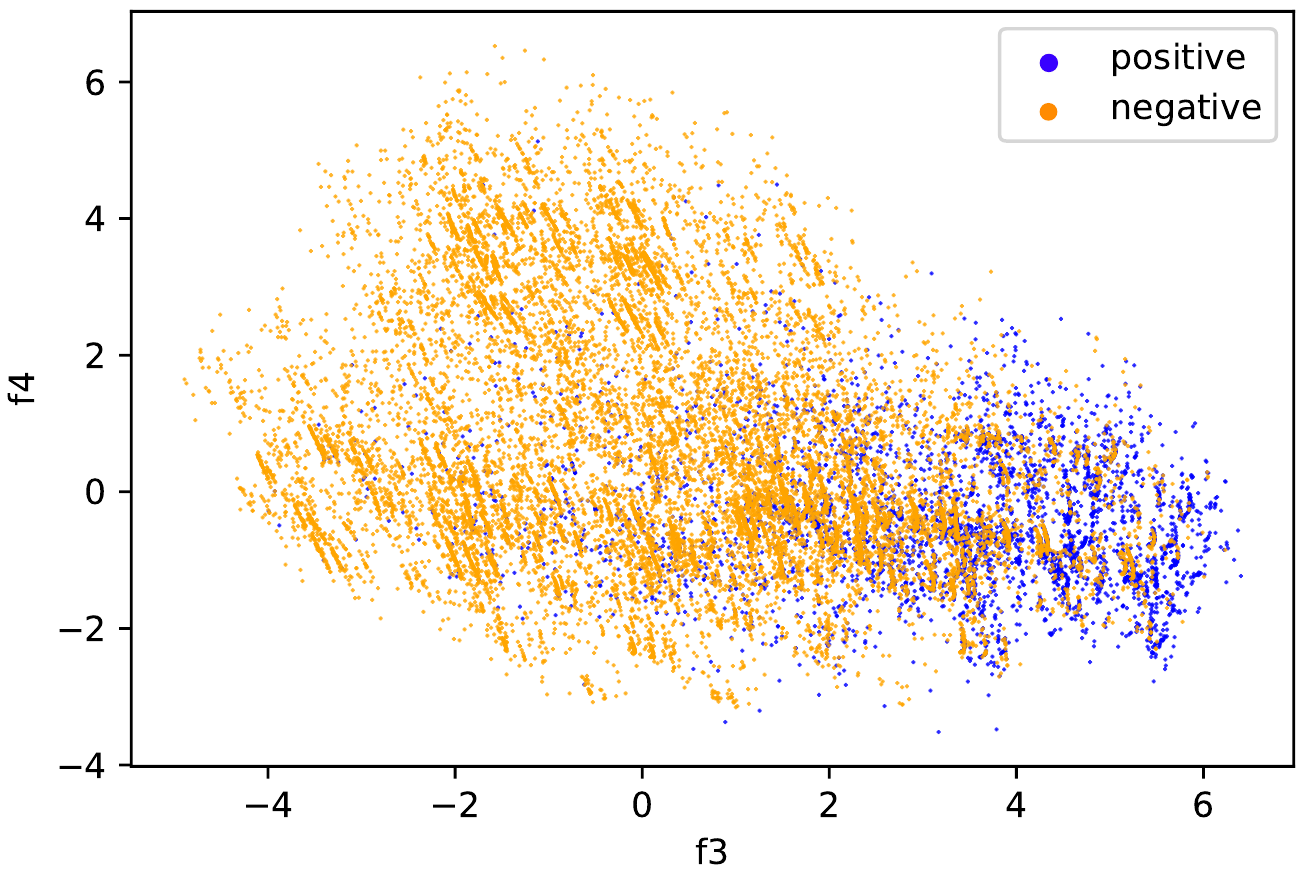}\\
		\end{minipage}
		\label{fig:visual_income}
	}
	\caption{Visualization of learned features colored by (a) gender and  (b) income. }
	\label{fig:visual}
\end{figure}

\subsection{Feature visualization}
\label{subsec:visual}
To better understand what kind of features are learned from the auto-encoder part, we visualize the extracted features by randomly selecting 2 dimensions of the 10 dimensional latent space
and color them according the protected attribute  (Fig.~\ref{fig:visual_gender}) and by the label (Fig.~\ref{fig:visual_income}) respectively. Fig.~\ref{fig:visual_gender} illustrates that the protected attribute information is mixed up in the latent space, which indicates that the fair representation is learned. Fig.~\ref{fig:visual_income} shows that the label information is distinguishable. The learned representation is not only fair but also suitable for the binary classification task which follows afterwards.

\subsection{The effect of the multi-loss function on the accuracy}
\label{subsec: ep_multiLoss}
In this experiment we evaluate the effect of the multi-loss function on the accuracy and compare the auto-encoder with MSE loss + Cross Entropy loss(we call the network AE-M), to an auto-encoder with normal MSE loss(AE-N). We set $\alpha=0, \beta=0$, which means to ignore the equations \ref{eq:ae_loss}  and \ref{eq:mlp_loss}.

\begin{figure}[!t]
	\centering
	\subfigure[Testing accuracy of AE-M]{
		\includegraphics[width = 5.42cm, height = 3.06cm]{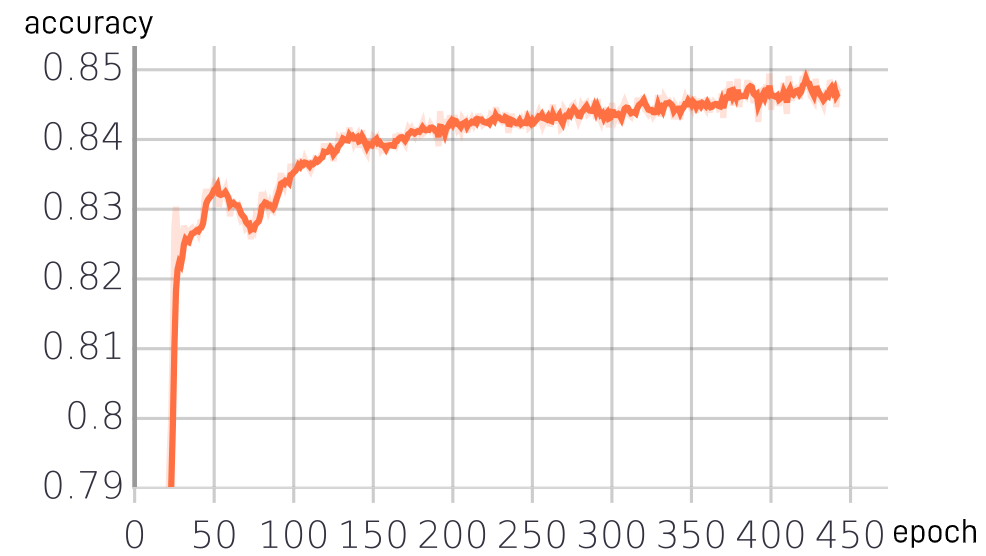}
		\label{fig:AE-M_acc}
	}
	\subfigure[Testing accuracy of AE-N]{
		\includegraphics[width = 5.42cm, height = 3.06cm]{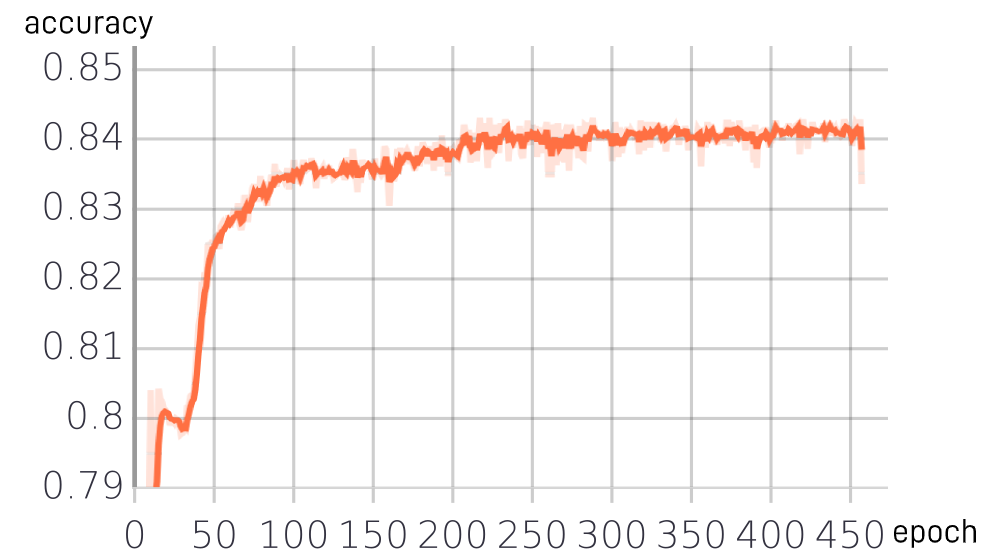}
		\label{fig:AE-N_acc}
	}	
	\caption{Comparison of auto-encoder with MSE+Cross-Entropy loss (left) and with normal MSE loss function (right).}
	\label{fig:mul_nor_loss}
\end{figure}

By observing the testing accuracy shown in Fig.~\ref{fig:AE-M_acc} and Fig.~\ref{fig:AE-N_acc}, we can conclude that AE-M does not perform worse but even achieves a slightly better predictive performance  (testing accuracy is $0.76\%$ 
higher than AE-N). 

\section{Conclusion}
\label{sec:conclusions}
In this work we propose \ourmethod, a neural network that performs joint feature representation and classification for fairness-aware learning. The network consists of two parts, an autoencoder part for representation learning and a classification part for fair decision making.
Our approach optimizes a multi-objective loss function which  (a) learns a fair representation  by suppressing protected attributes   (b) maintains the information content  by minimizing a reconstruction loss and  (c) allows for solving a classification task in a fair manner  by minimizing the classification error and respecting the equalized odds-based fairness regularizer. 
Our experiments demonstrate that such a joint approach is superior to a separate treatment of unfairness in representation learning or classifier learning. Our method achieves the highest accuracy and balanced accuracy rates.
The ablation studies in the experiments demonstrate the importance of each component we propose in our framework. 

Note that our architecture contains a branch of an auto-encoder which allows unsupervised learning. Thus, our framework is suited for semi-supervised learning with sparsely labeled data. We will elaborate on this aspect in future works.

%
%
%
\bibliographystyle{splncs04}
\bibliography{refs}
\end{document}

%% file: introduction.tex
The wide usage of AI-based systems, mostly powered nowaydays by data and machine learning algorithms, in areas of high societal impact raises a lot of concerns regarding accountability, fairness, and transparency~\cite{united2014big} of their decisions. 
Such systems can become discriminatory towards groups of people or individuals based on \emph{protected attributes} like gender, race, religious beliefs etc, as it has been already showcased in a variety of cases~\cite{AmazonPrime,datta2015automated,airbnb,sweeney2013discrimination,larson2016we}.
For example, \cite{datta2015automated} shows that Google’s ad-targeting system was displaying more highly paid jobs to men than to women, thus making discriminatory decisions based on gender.
Such incidents call for methods that explicitly target bias and discrimination in AI-systems, while maintaining their predictive power. The ever increased interest in this area is already reflected in the 
large, given the recency of the field, body of literature on fairness-aware learning and responsible AI, in general (see~\cite{doi:10.1002/widm.1356} for a recent survey).

However, despite the large number of methods and approaches for fairness-aware machine learning proposed thus far, most of these approaches refer to supervised learning upon a given feature representation. Some approaches that target fair representation learning also exist, e.g., \cite{DBLP:conf/nips/SamadiTMSV18} but they focus on learning a fair lower dimensional representation of the data which can be used either as a standalone result (e.g., for visualization purposes), or as an input to some other learning task (e.g., for learning a classifier upon the reduced representation). Only few approaches exist 
that jointly target fairness in both representation learning \emph{and} supervised learning, e.g.,~\cite{DBLP:journals/corr/EdwardsS15,madras2018learning}.

In this work we argue that a joint tackling of fairness in the machine learning pipeline (data $\rightarrow$ algorithm $\rightarrow$ model) is superior to the separate treatment of unfairness in representation- or supervised- learning. This is because bias-related corrections in representation learning do not guarantee that a model derived from the corrected data will be fair. Instead, the learning algorithm might still pick up certain data peculiarities that lead to discriminatory outcomes. Therefore, a joint \emph{goal-oriented} consideration in the pipeline  is much more effective, as also demonstrated in our experimental results.
To this end, we aim for a fair representation learning that preserves as much as possible the original data while obfuscating information on the protected attribute so decisions based on the protected attribute in the latent space are not possible. Additionally, the learned representation should structure itself in such a fashion, that a task-goal, such as a classification task, can still be appropriately solved.

 The aforementioned goals are implemented in our proposed \ourmethod method via a neural network with a jointly-optimized multi-objective loss function. In particular, the loss function aims at learning a fair representation (by suppressing protected attributes) that  maintains the information content (by minimizing the reconstruction loss) and allows for solving a classification task in a fair manner by minimizing the classification error and respecting a fairness-related regularizer. In this work we employ equalized odds as our fairness notion. 
Our experiments against several state of the art fairness-aware learning approaches demonstrate superior or highly competitive performance.

Our contributions can be summarized as follows:
\noindent
\begin{itemize}
\item We propose a neural network that learns a fair representation and a fair classifier \emph{jointly} in an end-to-end manner.
\item The contribution of the different components during training can be adjusted, leading to a very flexible and competitive framework. 
\item Our experiments demonstrate that \ourmethod with a goal-oriented fair representation is superior to a plain fair classifier without explicit representation constraints as well as to a standard fair representation learner without an explicit classification goal.
\item The source code will be made available (after acceptance).
\end{itemize}


The rest of the paper is organized as follows: Related work is summarized in Section~\ref{sec:related}. Necessary background is provided in 
Section~\ref{sec:basics}. Our joint goal-oriented approach to fairness-aware learning is introduced in Section~\ref{sec:method}. Experimental results are presented in Section~\ref{sec:experiment}. Finally, Section~\ref{sec:conclusions} concludes our work and identifies interesting directions for future research.

%% file: related.tex
Despite its recency, the domain of fairness-aware machine learning features already a rich variety of methods from fairness formalization to methods for bias discovery and mitigation. The latter can be further categorized into pre-processing, in-processing and post-processing approaches to fairness depending on whether they focus on mitigating discrimination at the data, algorithms or model output, respectively. 
In what follows, we provide an overview of the methods, focusing on the most relevant ones for our work.


\noindent\textbf{Formalizing fairness}:  
At least 20 different fairness notions have been proposed in the recent years only in the computer science domain 
~\cite{DBLP:journals/ker/RomeiR14,DBLP:conf/icse/VermaR18,DBLP:journals/datamine/Zliobaite17} but still there is an ongoing debate on the pros and cons of 
popular mathematical formalization of fairness and even on whether fairness can be boiled down to a mathematical equation.
Existing fairness notions can be categorized as follows~\cite{DBLP:conf/icse/VermaR18}: 
i) causal reasoning notions that
aim to detect hidden relationships among the attributes and outcomes based on directed acyclic graphs;
ii) predicted outcome notions that rely solely on algorithmic predictions - popular notions in this category include  statistical parity and $p$-rule.
iii) predicted and actual outcome notions that extend category (ii) by also taking into account the ground truth labels - popular notions in this category include equal opportunity and equalized odds~\cite{DBLP:conf/nips/HardtPNS16};
iv) predicted probabilities and actual outcome notions that extend category (iii) but instead of the predicted labels they employ the predicted probabilities and therefore they can be used for models with probabilistic outputs; 
v) similarity based methods that assume that ``similar" individuals should receive the same decision independent of their protected values - a popular notion in this category is  fairness through awareness~\cite{DBLP:conf/innovations/DworkHPRZ12}. 

\noindent\textbf{Mitigating fairness in supervised learning}: 
\textit{Pre-processing approaches to fairness} assume that there exist encoded (e.g., societal) biases in the data which they try to eliminate before ``feeding'' the  data to some learning algorithm. 
For example,~\cite{DBLP:journals/kais/KamiranC11} proposes instance re-weighting, label swapping, and data augmentation to eliminate discrimination in the input data. 
Similar ideas, but for the online scenario, were proposed by~\cite{DBLP:conf/dexa/IosifidisTN19}. Data augmentation has also been used in~\cite{iosifidis2018dealing} in order to force the model so as to learn efficiently all the population segments. 
In~\cite{iosifidis2019fae} a bagging schema is proposed to equalize the data distributions for the different population segments. 
In~\cite{DBLP:conf/nips/CalmonWVRV17} a probabilistic framework 
for discrimination-preventing preprocessing in supervised learning is introduced with the goal to preserve the utility of the data for the learning task while controlling the correlation between the protected attributes and class and minimizing instance distortion.
%
\textit{In-processing approaches to fairness} aim to explicitly consider fairness into the learning algorithm 
by constraining or regularizing the model during the training phase. It comprises the most popular category to fairness mitigation, which however depends on the algorithm per se.
For example, in~\cite{DBLP:conf/www/ZafarVGG17} the authors tweak the objective function of the linear SVM and Logistic Regression models by inserting 
convex-concave fairness-related constraints (they use equalized odds as fairness measure). 
In~\cite{DBLP:conf/icdm/KamiranCP10}, a fairness-aware splitting criterion for decision trees is proposed that evaluates not only the splitting quality w.r.t. the class but also the discrimination effect of a potential split. The work is extended in ~\cite{DBLP:conf/ijcai/ZhangN19} for online learning, using Hoeffding Trees as the underlying model. In~\cite{DBLP:conf/cikm/IosifidisN19} the authors aim to eliminate discrimination in sequential learning scenarios (in particular, boosting) by dynamically adapting the data distributions over the training rounds using a cumulative version of equalized odds. 
In~\cite{DBLP:conf/www/KrasanakisXPK18} it is assumed that there exist latent fair class labels (non-observable) which are estimated via an iterative process.
Finally, \textit{post-processing approaches to fairness} work directly at the output of a model and change its outcomes until a chosen fairness notion is satisfied. 
For example, ~\cite{DBLP:conf/sdm/FishKL16} shifts the decision boundary of AdaBoost w.r.t a protected attribute until statistical parity is achieved. In~\cite{DBLP:conf/nips/HardtPNS16} different thresholds are introduced for different population segments to enforce equal error rates. In~\cite{DBLP:journals/isci/KamiranMKZ18} the predictions of probabilistic classifiers and ensemble models for instances close to the decision boundary are altered until statistical parity is fulfilled. 
Our \ourmethod~belongs to the category of in-processing approaches as the objective function of the NN is altered to account for fairness. In contrast to the majority of the previous approaches however, our method comprises a \emph{joint} approach for fair-feature representation- and classifier-learning. 

\noindent\textbf{Fair representation learning approaches}: 
Fair representation learning aims to learn a transformation to a lower dimensional space where the protected and non-protected groups are indistinguishable. In~\cite{DBLP:conf/nips/SamadiTMSV18} the authors propose Fair-PCA, an extension of PCA, that forces similar reconstruction errors between protected and non-protected groups. 
In~\cite{DBLP:journals/corr/LouizosSLWZ15}, the Variational Fair Auto Encoder is proposed that is able to also learn fair non-linear functions, which can be used after as input to other learning models. Our \ourmethod also derives non-linear transformations via autoencoders, however on the contrary to~\cite{DBLP:journals/corr/LouizosSLWZ15}, we dont only focus on fair-representation learning but rather on joint representation-and classifier-learning. 
In~\cite{DBLP:journals/corr/abs-2002-10312} an approach for learning individually fair representations is proposed using an end-to-end model with autoencoders. On the contrary, our~\ourmethod aims at learning representations that are fair for each group (i.e., protected and non-protected).

Closer to our work are the joint approaches~\cite{DBLP:journals/corr/EdwardsS15,madras2018learning} that aim at both fair representation- and classifier-learning.
In~\cite{DBLP:journals/corr/EdwardsS15,madras2018learning} instead of using some constraining to reduce the dependencies on the sensitive attribute in the latent space (e.g., by minimizing KL-divergence as in our \ourmethod), they train an adversary classifier to discriminate between the protected and non-protected groups. In particular, in~\cite{DBLP:journals/corr/EdwardsS15} they optimize for statistical parity, whereas \cite{madras2018learning} extends the idea for more fairness measures.
It is not clear in what circumstances a constraint-based approach or an adversary one should be preferred~\cite{DBLP:journals/corr/EdwardsS15}, but we include~\cite{madras2018learning} in our experimental analysis.




%% file: basics.tex
Let $A = \{A_1,...,A_d\}$ be a $d$-dimensional feature space of mixed attribute types. We assume the existence of a protected attribute $S \in A$, e.g., $S = gender$. We assume $S$ is binary: $S = \{s, \bar{s}\}$,  with $s$ denoting the protected group (e.g., $s=female$), and $\bar{s}$ the non-protected group e.g., $\bar{s}=male$.
An instance $X \in A_1 \times A_2  \cdots \times A_n$ is a $d$-dimensional feature vector representing an object in the vector space $A$. Each instance is assigned a label $c \in C$ by some unknown target function $g: A \rightarrow C$. 
For simplicity, we assume the class attribute is also binary, i.e., $C=\{+,-\}$. 
We use the notation $s_+$ ($s_{-}$), $\bar{s}_+$ ($\bar{s}_{-}$) to denote the protected and non-protected group for the positive (negative, respectively) class.

The target function $g()$ is unknown, instead a training set $D=\{(X_i,c)\}$ of i.i.d. instances drawn from the joint attribute-class space $A \times C$ is available and can be used for approximating $g()$. 
The goal of fairness-aware supervised learning is to approximate $g()$ via a mapping function $f()$ that does not only map correctly future unseen instances of the population from $A$ into $C$, but also mitigates discriminatory outcomes.
The former aspect corresponds to the typical objective of supervised learning achieved through empirical risk minimization. The latter aspect is evaluated in terms of some fairness measure (c.f.  Section~\ref{sec:related}). 

\subsection{Formalizing fairness}

In this work, we employ Equalized Odds~\cite{DBLP:conf/nips/HardtPNS16} (shortly \textit{Eq.Odds}) as our fairness measure. $Eq.Odds$ accounts for the percentage difference among protected and non-protected groups in the model's outcomes. In particular, let $\delta FPR$ ($\delta FNR$) be the difference in false positive rates (false negative rates, respectively) between the protected and non-protected groups, defined as follows:
\begin{equation}
\label{eq:FPR_FNR}
\begin{split}
\delta FPR = P(c\neq \dot{c} | \bar{s}_-) - P(c\neq \dot{c} | s_-) \\
\delta FNR = P(c\neq \dot{c} | \bar{s}_+) - P(c\neq \dot{c} | s_+) 
\end{split}
\end{equation}
where $\dot{c}$ are the predicted labels. The goal of $Eq.Odds$ is to minimize both differences:
\begin{equation}
Eq.Odds = |\delta FPR |+ |\delta FNR|
\label{eq:EqOdds}
\end{equation}
where $Eq.Odds \in [0,2]$, with $0$ indicating no discrimination and 2 indicating maximum discrimination. 

$Eq.Odds$ has become quite popular among recent state-of-the-art fairness-aware methods~\cite{DBLP:conf/cikm/IosifidisN19,DBLP:conf/nips/HardtPNS16,DBLP:conf/www/ZafarVGG17,DBLP:conf/www/KrasanakisXPK18,madras2018learning}.
In contrast to the well-known statistical parity~\cite{DBLP:journals/kais/KamiranC11}, which uses only the positive predicted outcomes without the aid of true labels, or equal opportunity~\cite{DBLP:conf/nips/HardtPNS16}, which accounts only for the false negative difference among $s$ and $\bar{s}$, $Eq.Odds$ is able to locate discriminatory outcomes for both classes. Furthermore,
statistical parity is prone to favor groups by discriminating on specific individuals~\cite{DBLP:conf/innovations/DworkHPRZ12}.

\subsection{Auto-encoders} 
\label{subsec:ae}

An auto-encoder (AE) is an unsupervised neural network that learns an approximation of the identity function such that the output of the network is similar to its input. 
A reduced/compressed representation is learned by placing constraints in the structure of the network, e.g. by using a bottleneck layer. 

In this work, we consider
mixed attribute type data of numerical and nominal attributes.
%
Reconstructing the numerical attributes could be considered as a regression task, so we use the \textit{Mean Square Error} as the loss function for numerical attributes. Since for the nominal attributes there is no order among their values, 
reconstructing their values could be considered as a classification task, so we use the \textit{Cross Entropy} as the loss function for nominal attributes. We assume there exist $K$ numerical and $N$ nominal features, such that $K+N=d$.
We combine the feature-type specific loss functions in the overall objective function of the auto-encoder as follows (we compute the loss per batch of $B$ instances):
\begin{equation}
     L\left ( X,\hat{X} \right )=\frac{1}{B}\sum_{b=1}^{B}\left (\sum_{k=1}^{K}\left ( X_{b,k} - \hat{X}_{b,k} \right )^{2} - \sum_{j=1}^{N} \sum_{l_{j}=1}^{M_{j}} X_{b,l_j} \log\left ( p_{b,l_j} \right )  \right )
	\label{eq:multi_loss}
\end{equation} 
where $X$ is the original instance, $\hat{X}$ is the reconstructed instance and $X_{ji}$ is the value of instance $j$ in dimension $i$.  
The first term of the above equation refers to the loss of numerical attributes:  
$X_{b,k}$, $\hat{X}_{b,k}$ denotes the original and reconstructed data of numerical attributes, respectively. 
The second term of the the above equation  refers to the loss of nominal attributes.
For each nominal attribute $j$, $l_{j}$ represents the class label and $M_{j}$ the number of values of the feature.
For the $j$-th nominal attribute in instance $b$, $X_{b,l_{j}}$ has the binary value (positive or negative) which indicates if the class label $l_{j}$ is the correct classification, $p_{b,l_{j}}$ represents the predicted probability of class $l_{j}$.
